\documentclass[runningheads]{llncs}

\usepackage[T1]{fontenc}

\usepackage{graphicx}
\usepackage{subfigure}
\usepackage{tabularx}
\usepackage{mathtools}
\usepackage{subcaption}
\usepackage{orcidlink}
\usepackage{amssymb}
\usepackage{natbib}
\title{Modeling Sustainable Resource Management using Active Inference}

\author{
  Mahault Albarracin\inst{1}\orcidlink{0000-0001-2345-6789} \and
  Ines Hipolito\inst{2}\orcidlink{0000-0002-3456-7890} \and
  Maria Raffa\inst{3}\orcidlink{0000-0003-4567-8901} \and
  Paul Kinghorn\inst{4}\orcidlink{0000-0004-5678-9012}
}

\authorrunning{Albarracin et al.}

\institute{
  Université du Québec à Montréal, Montréal, Canada \and
  Macquarie University, Sydney, Australia \and
  IULM University, Milan, Italy \and
  Department of Informatics and Engineering, University of Sussex, Brighton, UK
}

\begin{document}

\begin{center}
    {\Large \bfseries \boldmath Modeling Sustainable Resource Management using Active Inference \par}
    \vspace{1cm}
    {\large Mahault Albarracin$^{1}$\orcidlink{0000-0001-2345-6789}, 
    Ines Hipolito$^{2}$\orcidlink{0000-0002-3456-7890}, 
    Maria Raffa$^{3}$\orcidlink{0000-0003-4567-8901}, 
    Paul Kinghorn$^{4}$\orcidlink{0000-0004-5678-9012} \par}
    \vspace{0.5cm}
    {\small 
        $^{1}$Université du Québec à Montréal, Montréal, Canada \\
        $^{2}$Macquarie University, Sydney, Australia \\
        $^{3}$IULM University, Milan, Italy \\
        $^{4}$Department of Informatics and Engineering, University of Sussex, Brighton, UK
    }
\end{center}

\begin{abstract}
Active inference helps us simulate adaptive behavior and decision-making in biological and artificial agents. Building on our previous work exploring the relationship between active inference, well-being, resilience, and sustainability, we present a computational model of an agent learning sustainable resource management strategies in both static and dynamic environments. The agent's behavior emerges from optimizing its own well-being, represented by prior preferences, subject to beliefs about environmental dynamics. In a static environment, the agent learns to consistently consume resources to satisfy its needs. In a dynamic environment where resources deplete and replenish based on the agent's actions, the agent adapts its behavior to balance immediate needs with long-term resource availability. This demonstrates how active inference can give rise to sustainable and resilient behaviors in the face of changing environmental conditions. We discuss the implications of our model, its limitations, and suggest future directions for integrating more complex agent-environment interactions. Our work highlights active inference's potential for understanding and shaping sustainable behaviors.

\end{abstract}
\keywords{Active Inference \and Sustainability \and Generative model.}

\section{Introduction}

For the past decade, we have shown that the Free Energy Principle can serve as a foundational concept in  predicting and modeling the present and future behaviors of a system. Under this principle, the behavior of a system aims to maintain equilibrium and sustaining life through the minimization of free energy \citep{parr2022, parr2019, stubbs2024}. Systems, particularly biological ones, act to minimize the difference between their representation of the world, encoded in internal states and the external environment. By reducing this discrepancy, as quantified by free energy, the system achieves a state of balance and effectively adapts to its surroundings \citep{friston2017, friston2023a, parr2023a}. 
We can thus understand it as a measure of uncertainty or surprise, used such that agents are driven to more predictable and stable states. Using free energy, systems can slowly make adjustments and adaptations, and thus maintain homeostasis in a changing environment \citep{ramstead2018, kirchhoff2018, karl2012, dacosta2023a, pezzulo2024a}. 
This modeling approach has been used for various types of systems, from neural processes and cognitive functions to broader ecological and social dynamics \citep{friston2010, dacosta2024, solymosi2024,albarracin2024a, matsumura2023, montgomery2023, ramstead2020, pezzulo2024a}.

While it is a widely held assumption that all systems will invariably minimize free energy (FE), this is not always a simple linear process. To understand this, we have to consider the system's goals and constraints. These goals can sometimes result in behaviors that do not align perfectly with immediate free energy minimization. This is partly what can make a system, given a specific scale of measurement, somewhat unsustainable.
Not all systems are capable of effectively minimizing free energy.  They may indeed have constraints in their structure or function. Think of certain pathological conditions impeding a system's ability to minimize free energy efficiently. These conditions can lead to maladaptive behaviors or states that deviate significantly from what would be predicted by the FEP. For example, constraints in structure or function: in neurological disorders such as schizophrenia, the brain's ability to minimize free energy can be impaired. As it distorts connectivity, it may also alter perceptions and thoughts - no longer fully related to the external world. Someone with Schizophrenia can struggle to reduce uncertainty about its environment, resulting in maladaptive behaviors \citep{friston2016, harikumar2023, zarghami2023}.

Free energy minimization can also be influenced by external perturbations and environmental factors. The very nature of the environment is unpredictable dynamics, which temporarily disrupt a system's meta-stable states. The system transiently increases free energy as it adapts to new conditions. Albarracin et al. (2024) explore how systems must deal with external shocks and stresses (perturbations) to maintain sustainability, resilience and well-being. They suggest that resilience means absorbing shocks and stresses from the environment, while sustainability requires enduring capacity to stay resilient, but without causing a loss of resilience of the environment super-system. In this paradigm, external perturbations are central to developing better strategies to maintain well-being across system strata.
Since these perturbations can be unpredictable, the temporality of strategies can change. Long-term strategies can weather slight increases in free energy temporarily to achieve more stable and favorable conditions in the future.  This is the case we will be testing and presenting in this paper: long-term strategies involving temporary increases in free energy. When an  agent learns that it does not have to satisfy its greed immediately, even if it is very hungry, because the aim is to maintain a balance between itself and the environment (such as a room with food) over time, the agent can resist the urge for immediate gratification and managing its resources judiciously. And thus, the agent can endure short-term discomfort (increased free energy) to ensure long-term stability and sustainability. 

To do so, we test out two cases, detailed in the Methods section. Case 1 acts as a baseline scenario, and involves a static environment where the agent decides whether to eat food or not. 
In Case 2, the environment is dynamic, and the agent must learn to moderate its consumption behavior over time. Food increases when the agent does not eat, introducing a dynamic aspect to the environment. 
This study is important for two reasons. First, we must understand adaptive strategies to properly predict when systems will achieve long-term stability and sustainability. It will help us predict when systems choose to balance short-term needs with long-term goals. But this will also help us identify potential vulnerabilities, such that we can identify areas where intervention may be needed to prevent collapse or dysfunction.
The FEP dictates that behavior should align with minimizing free energy. But we have to better understand the variability in paths where this principle isn't consistently upheld at a given scale of measurement.

\section{Methods}

\renewcommand{\thefigure}{1}
\begin{figure}[htbp]
\centering
\includegraphics[width=\textwidth]{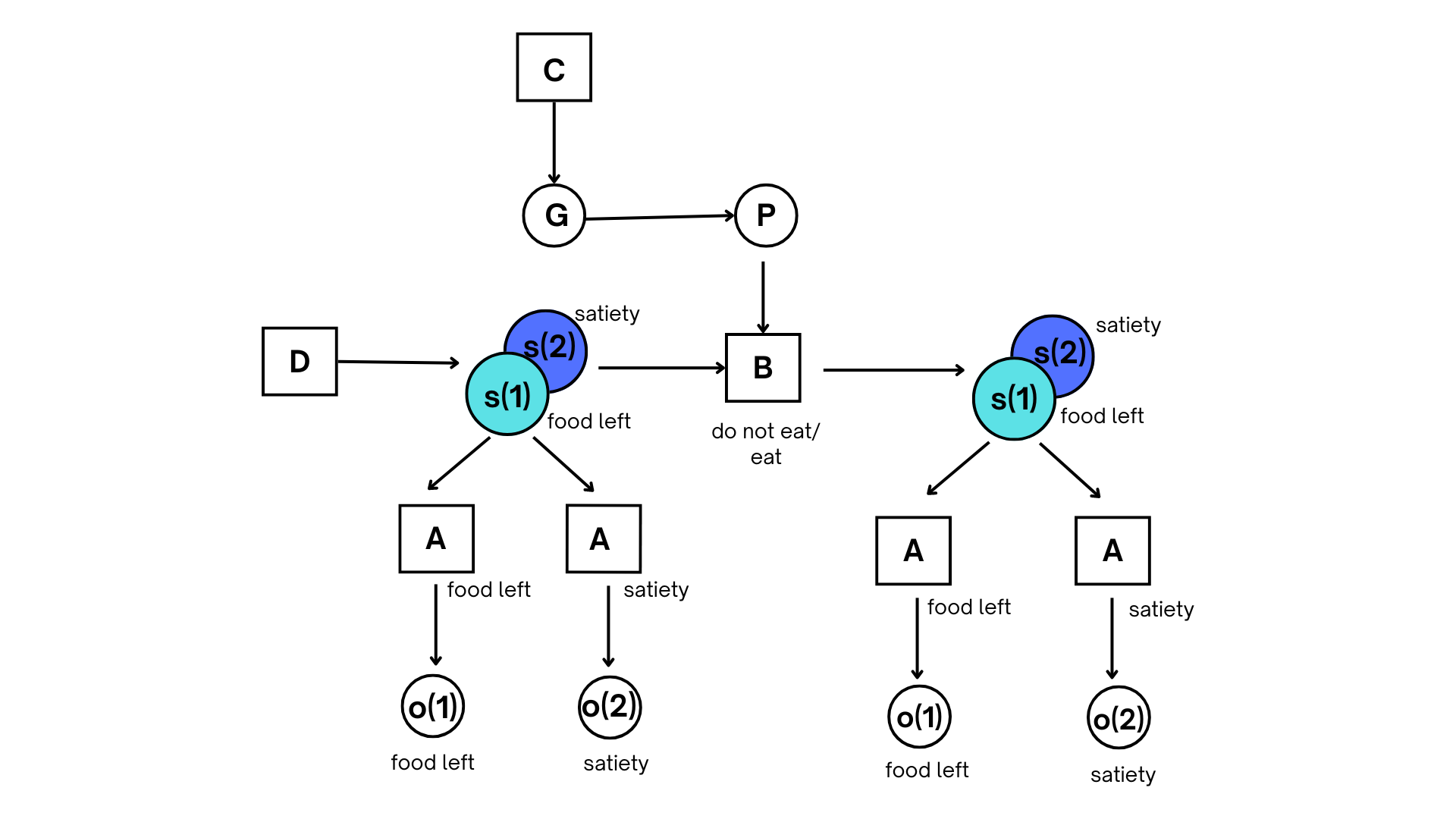}
\caption{The agent’s generative model encodes beliefs about the causal structure of the environment and how its actions affect the state of the world. The true state of the environment is represented by two hidden state factors - the availability of food
(s1) and the agent’s satiety (s2). The prior preference C matrix specifies the agent’s innate drives or goals, in this case a strong preference for being satiated. The starting conditions are specified by the initial state distribution, D. Here, food is initially present
but the agent is not satiated. The agent has two observation modalities - the presence of food (o1) and its own satiety level (o2). The agent can select between two actions at each time step - "eat" or "do not eat". We have two hidden state factors: food left and satiety. For the "do not eat" action, for Case 1 the B matrix is an identity matrix, as this action does not change the state, while for Case 2 it changes, since not eating leads to an increase in available food. When the agent chooses the "eat" action, if food is present, the states will transition by reducing "food left" by 1 (down to a minimum of 0) state and by increasing "satiety" by 1 (to a maximum of 2).}

\label{fig:image2}
\end{figure}

\subsection{Case 1: Static environment}
We build this simulation using the PyMDP package, by Conor Heins and colleagues \citep{heins2022pymdp}. In Case 1, we consider a static environment where the agent's goal is to maintain satiety by deciding whether to eat the available food. The generative model for Case 1 (detailed in Table 1, and visualized in figure 1) includes hidden states for food availability and agent satiety, observations that directly correspond to these hidden states, and actions to eat or not eat.
The Likelihood Matrix (\(\mathbf{A}\)) assumes an identity mapping between hidden states and observations, meaning that the agent directly observes the true states of food availability and its own satiety. Mathematically, this is represented as \( P(o_t \mid s_t, \mathbf{A}) = \text{Cat}(\mathbf{A}) \), where \(\mathbf{A}\) is an identity matrix. This means the probability of an observation given a state is 1 if they correspond and 0 otherwise:

\[
P(o_t = i \mid s_t = j) =
\begin{cases} 
1 & \text{if } i = j \\
0 & \text{if } i \neq j 
\end{cases}
\]

The agent performs variational inference to optimize an approximate posterior \( Q(s_t) \) over hidden states at each timestep, using the expected log likelihood of observations \( \mathbb{E}[\log P(o_t \mid s_t)] = Q(s_t)^T \log \mathbf{A} \).

The Transition Matrix ($B$) specifies that the "eat" action leads to satiety when food is present, while food remains constantly available regardless of the agent's actions. The "don't eat" action leads to hunger.The transition likelihood $B$ is represented as a set of matrices $B[f]$, one for each hidden state factor $f$, with dimensions $S_f \times S_f \times U_f$, where $S_f$ is the number of levels for factor $f$ and $U_f$ is the number of control states or actions for that factor.

The entry $B[f][i, j, k]$ represents the probability of transitioning from state $j$ to state $i$ for factor $f$, given action $k$: $P(s_{t+1}^f = i \mid s_t^f = j, u_t^f = k)$. In this case, the "eat" action ($k = 0$) would have a high probability of leading to the "satiated" state ($i$) when the current state is "food available" ($j$), while the "don't eat" action ($k = 1$) would likely lead to the "hungry" state.

Importantly, the transition matrices $B[f]$ are assumed to be conditionally independent across factors, meaning that the next state of factor $f$ only depends on the current state and action for that factor, and not on the states of other factors: $P(s_{t+1}^f \mid s_t^f, u_t^f) = P(s_{t+1}^f \mid s_t, u_t)$. This simplifies the computation of the joint transition probability.

The Preference Vector ($C$) encodes a strong preference to observe satiation and food present. The agent's goals and preferences are represented as a prior distribution over observations, $P(o_{1:T})$. The $C$ vector encodes these preferences as a categorical distribution, where higher values correspond to preferred observations. The agent aims to maximize the probability of sampling these preferred observations.

The Initial State Distribution ($D$) is not specified, so that it is a uniform distribution where each state has an equal probability of being the initial state. In the PyMDP framework, the initial state distribution is represented as a categorical distribution over hidden states at the first timestep, $P(s_1 \mid D) = \text{Cat}(D)$. If not specified, it defaults to a uniform distribution, assigning equal probability to all possible initial states.

During the generative process, the agent interacts with the environment, and its actions affect the state transitions according to the generative process. If the agent chooses not to eat, the state of the environment remains  unchanged. If the agent chooses to eat and food is present, the agent becomes satiated, but food remains available due to the static nature of the environment.
We then instantiate the simulation loop. First, the agent performs state inference based on the current observation, evaluating policies to maximize expected free energy, and selecting an action that minimizes free energy and aligns with its preferences. The selected action is applied to the environment, resulting in state transitions and new observations, and the loop continues with the agent updating its beliefs, inferring policies, and selecting actions to achieve its goals.
In the extension to Case 1, we introduce variations to test the agent's adaptability. In Case 1.1, we set incorrect A and B matrices, introducing flawed perceptions and beliefs about state transitions. This extension is intended for us to validate that the behavior of the agent is in fact predicated on its appropriate appraisal of the environment.


\begin{table}[ht]
\centering
\begin{tabular}{|m{4cm}|m{4cm}|m{4cm}|}
\hline
\textbf{Component} & \textbf{Values} & \textbf{Description} \\
\hline
\hline
Hidden States & Food availability: present (1), absent (0) & Represents the true state of food availability in the environment \\
& Agent satiety: hungry (0), satiated (1) & Represents the true state of the agent's satiety \\
\hline
Observations & Observed food availability & Corresponds directly to the food availability state \\
& Agent's perceived satiety & Corresponds directly to the agent satiety state \\
\hline
Actions & Eat (1), Don't Eat (0) & The actions available to the agent \\
\hline
Likelihood Matrix (A) & Identity mapping  & Assumes the agent directly observes the true states \\
\hline
Transition Matrix (B) & "Eat" (1) leads to satiety (1) when food is present (1) & Specifies the state transitions based on the agent's actions \\
& "Don't Eat" (0) leads to hunger (0) & \\
\hline
Preference Vector (C) & Strong preference for satiated (1) and food present (1) & Encodes the agent's goals and drives its behavior \\
\hline
Initial State Distribution (D) & Uniform Distribution& Sets the starting conditions for the simulation \\
\hline
\end{tabular}
\caption{Components of the generative model for Case 1}
\label{tab:case1_components}
\end{table}

\subsection{Case 2: Dynamic Environment}
In Case 2 (laid out in table 2), we extend our model to a dynamic environment where the agent's actions have consequences on the availability of food resources over time. The goal is to study how the agent adopts a sustainable behavior, balancing its immediate need for satiety with the long-term availability of food.
The generative model for Case 2 (Figure 1) builds upon the previous model by introducing more granularity in the states and observations, allowing for a wider range of behaviors and interactions between the agent and the environment. Both the observations and hidden states are expanded to have three levels each: food left (0: none, 1: some, 2: abundant) and satiety (0: not satiated, 1: somewhat satiated, 2: fully satiated).
In this model, we assume that the agent directly observes the true environmental states with some variations across different levels of food availability and satiety. In this dynamic environment, the transitions depend on both the current state and the action taken by the agent. If the agent does not eat, food availability increases over time, while if the agent eats, food availability decreases or remains depleted. For the satiety state, if the agent does not eat, satiety decreases over time, while if the agent eats, satiety increases.
In Case 2, the preferences are designed to balance between maintaining satiety and ensuring a sustainable food supply, encouraging the agent to maximize its satiety while also considering the long-term availability of food resources. Specifically, the agent has a strong preference for being satiated, while flat preference over food left.  The agent interacts with the dynamic environment over multiple time steps, updating its beliefs and actions based on the observed states and the changing dynamics of the environment, and it learns to not eat even if it is not fully satiated.  The agent is initialized with the generative model specified in Case 2, using an extended policy length to plan multiple time steps ahead and anticipate future consequences. 
In the simulation loop, the environment starts with food fully available and the agent being half satisfied. 
The agent can plan over multiple time steps (policy length of 3), and thus has the opportunity to balance immediate consumption with long-term sustainability. The extended policy length allows the agent to anticipate future states and avoid greedy behavior that could lead to starvation. The agent's policies are restricted to ensure consistent and sustainable actions across all time steps for both observation modalities.

\begin{table}[ht]
\centering
\begin{tabular}{|m{4cm}|m{4cm}|m{4cm}|}
\hline
\textbf{Component} & \textbf{Values} & \textbf{Description} \\
\hline
\hline
Hidden States & Food left: none (0), some (1), abundant (2) & Represents the true state of food availability in the environment \\
& Agent satiety: not satiated (0), somewhat satiated (1), fully satiated (2) & Represents the true state of the agent's satiety \\
\hline
Observations & Observed food availability & Corresponds to the food availability state with some variability \\
& Agent's perceived satiety & Corresponds to the agent satiety state with some variability\\
\hline
Actions & Eat (1), Don't Eat (0) & The actions available to the agent \\
\hline
Likelihood Matrix (A) & High probability of correct observations, lower for adjacent states & Defines the probability of observations given the true hidden states \\
\hline
Transition Matrix (B) & ``Eat'' (1): food left decreases, satiety increases & Specifies the state transitions based on the agent's actions and current state \\
& ``Don't Eat'' (0): food left increases, satiety decreases & \\
\hline
Preference Vector (C) & Strong preference for satiety. Balances maintaining satiety and sustainable food supply & Encodes the agent's goals and drives its behavior \\
\hline
Initial State Distribution (D) & Uniform Distribution & Sets the starting conditions for the simulation \\
\hline
Policy Length & 3 time steps & Allows the agent to plan ahead and consider long-term effects \\
\hline
\end{tabular}
\caption{Components of the generative model for Case 2}
\label{tab:case2_components}
\end{table}

In Case 2.1, we extend the dynamic environment setup from Case 2 by introducing a learning mechanism for the agent. The key change is that the agent now starts with a random B matrix and updates it based on its experiences in the environment. The agent starts with a random B matrix instead of a predefined one, which will be updated as the agent interacts with the environment. The B matrix is initially random, and the agent updates this matrix based on observed transitions between states. The A matrix remains the same as in Case 2, mapping the hidden states to observations, and the C vector, representing the agent's preferences over observations, remains unchanged from Case 2.
To learn, the agent starts with a randomly initialized B matrix, which does not initially capture the correct state transitions. The random initialization is done using a Dirichlet distribution to ensure valid probability values.
At each time step, the agent receives observations, infers states, infers policies, and samples actions, similar to Case 2.
After executing an action and receiving the next observation, the agent updates its B matrix. The agent notes the transition from the previous state to the current state given the action taken. The B matrix is updated using a learning rate to adjust the probabilities of the observed transitions. For states that depend on a single factor, the transition probability is updated directly by increasing the probability of the observed transition by the learning rate. For states that depend on two factors (e.g., satiety depends on both food left and previous satiety), the transition probabilities are updated based on the dependencies specified. After updating, the B matrix is normalized to ensure that the probabilities sum to 1, maintaining a valid probability distribution.
We also introduce several extended case variations for Case 2 (with and without learning) to explore the agent's behavior and performance under different conditions. We test the agent's robustness by initializing the B matrix with incorrect values set to very high (1) or very low (0). The agent's performance is expected to degrade, demonstrating the importance of accurate transition models, and avoiding inertia. 
We examine the agent's behavior when it has different prior preferences by modifying the C vector to represent a strong preference for food being present. The agent prioritizes actions that ensure food is present, potentially at the expense of satiety.
We test the agent in an environment where food increases at a slower rate (0.5 units per step, compared to 1 unit per step previously) when not eating and decreases at a faster rate (1 unit per step) when eating. Satiety decreases faster when not eating (0.2 units per step, compared to 1 previously) and increases at a different rate when eating (0.8 units per step, compared to 1 previously). The agent needs to adapt its strategy to account for these specific changes in the environment dynamics. Its performance may be lower compared to Case 2 due to the increased difficulty in balancing food and satiety levels, as the food depletes more quickly when eating and satiety decreases more rapidly when not eating.
We finally assess the impact of planning horizon on the agent's performance by comparing agents with different planning horizons (1 time step vs. 3 time steps). Agents with a longer planning horizon are expected to perform better, as they can anticipate future states more effectively and make decisions that lead to more sustainable resource management.

\section{Results}

In Case 1, the agent is in a static environment where food is always available, and its task is to maintain satiety by deciding whether to eat. The agent consistently chooses to eat at every time step, reflecting its understanding that food is always available and that eating maximizes its satiety (Appendix, figure 3, first row). Food availability remains constant throughout the simulation, as expected in a static environment where food does not deplete (Appendix, figure 3, second row). The agent's satiety increases as it eats and remains at a high level, indicating successful learning and adaptation to maintain its internal state optimally (Appendix, figure 3, third row). This setup demonstrates the agent's ability to perform optimally in an environment with constant resources.
In Case 1.1, we introduce errors in the A and B matrices to test the agent's resilience and adaptability when its internal model does not accurately represent the environment. The agent's actions show a more erratic pattern, reflecting confusion or uncertainty due to the incorrect matrices. Despite the confused matrices, food availability remains constant as in the standard case (Appendix, figure 4, second row). The agent's satiety fluctuates more compared to the standard case, indicating that the agent's ability to maintain a consistent internal state is impaired by the incorrect perception and planning models. This case shows how deviations from accurate environmental models can affect an agent's behavior and performance, leading to less optimal decisions.
With Case 1 results, we have shown that the agent has a degree of validity, and that it does in fact show how the agent reacts to model fitness, and chooses the best actions relative to its own survival.
In Case 2, the environment is dynamic, with food depleting when eaten and replenishing if not consumed. The agent must balance its eating behavior to avoid starvation and resource depletion. It is equipped with a strong preference over satiety = 2, and flat preference over food left (Appendix, figure 5). Over multiple runs with a policy length of 3 time steps, the agent tends to avoid eating, leading him to die of starvation (Appendix, figure 6 left), or eats too much, leading him to death as well (Appendix, figure 6 right) as his food gets depleted.
In Case 2 with learning, the agent starts with a randomly initialized B matrix and updates it through interactions with the dynamic environment. The agent's actions fluctuate regularly between "Eat" and "Do Not Eat," suggesting that it has learned a strategy to balance its actions, so that it manages to survive the whole time of the run and keeps satiety between 0 and 1 (Appendix, figure 7). The survival time plot for each run shows that the agent consistently survives for the maximum number of time steps after the initial learning phase, indicating that it quickly learns an effective strategy to avoid starvation and maintain survival (Figure 2).

\renewcommand{\thefigure}{2}
\begin{figure}[htbp]
\centering
\includegraphics[width=\textwidth]{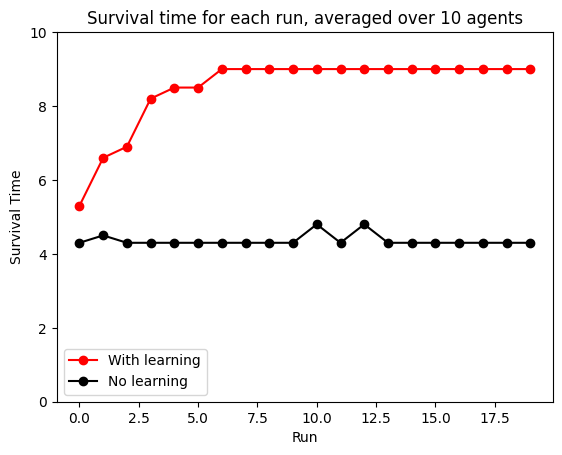}
\caption{Case 2, showing the survival time of agents with both learning and no learning when the agent starts with a random B matrix. The survival time plot for each run averaged over 10 agents, shows that the agents can quickly learn to survive by acting in a sustainable way.}

\label{fig:image9}
\end{figure}

Compared to Case 2 without learning, the case with learning shows the agent's ability to learn from its interactions with the environment and develop more effective strategies for survival and resource management.
The extended Case 2 variations explore the agent's behavior and performance under different conditions, such as incorrect transition models, altered prior preferences, and varying planning horizons. With an incorrect B matrix, where the values are set to extremes from the start (1 and 0, rather than lower probabilities), the agent consistently chooses to eat in every time step, leading to suboptimal behavior and eventual starvation (Appendix, figure 8, left). Under certain conditions, the agent was unable to learn, being stuck in the inertia of its transitions. But overall, with learning enabled, the agent was able to pull itself out of high values and was able to survive - highlighting the value of plasticity to get out of bad bootstraps.

In the case of strong preferences on both states (high satiety and high food left - without learning), the agent initially chooses to eat but then stops eating as food becomes scarce, demonstrating the influence of strong preferences on the agent's actions (Appendix, figure 9, left). This leads the agent to die over most of its runs quite quickly. With learning enabled (Appendix, figure 9, right), the agent is able to balance its actions again and can survive longer, balancing its preferences and the environmental demands.

When the environment rate of change changes (Appendix, figure 10), the agent's performance declines compared to the previous case, but learning still provides a significant advantage. With learning enabled, the agent adapts its strategy - eating less often to conserve food, maintaining higher average food levels, and sustaining satiety more effectively. This allows the agent to consistently survive the full run timesteps when learning, while it only survives around 3 timesteps without learning. Although the tougher environment dynamics make it more challenging, as the agent must plan in a different way and possibly over longer timescales, the agent demonstrates an impressive ability to adjust its policy through learning to match the new rate of change in food and satiety levels. Learning is critical for the agent to find the right balance and survive in this more complex scenario.

With a policy length of 1, the agent does very poorly without learning, dying basically after the first step. With learning, it takes the agent a little bit of time to learn how to survive, but it eventually does. Its actions are a little more erratic, but it does find a short term strategy (Appendix, figure 11). However, even with this short term strategy, it is unable to survive for very long, truly highlighting the need to focus on longer term strategies. 

\section{Discussion}
Our sustainable agent demonstrates how active inference can give rise to sustainable resource management strategies at the level of an individual agent. The agent's behavior emerges from the interaction between its model of the world, prior preferences, and the environmental dynamics. It seeks to optimize for immediate needs (e.g., hunger) and long-term outcomes (e.g., consistent food availability), learning to balance consumption and resource replenishment to promote sustainability.
Our findings align with our previous formalization of sustainability, resilience, and well-being within the active inference framework.

In Case 1, the static environment allowed the agent to exhibit inertia, maintaining a consistent consumption pattern without considering long-term resource availability. While this behavior was adequate for the given context, it lacked the flexibility needed for sustainable outcomes in more dynamic environments.
Case 2 introduced environmental variability, requiring the agent to demonstrate elasticity and plasticity. The agent's ability to adapt its eating habits in response to changing food availability exemplifies elasticity, as it temporarily endures increases in free energy (i.e., hunger) to ensure long-term stability. The agent's capacity to learn and update its model of the world based on new information reflects plasticity, enhancing its resilience in the face of environmental shifts.
The agent's adaptive behavior in Case 2 reflects resilience, as it adjusts its actions to maintain well-being under changing resource availability. The dynamic coupling between agent and environment in the study of sustainable resource management was critical, even at the level of a single agent. In the extended cases, we can see the issues with inertia, and the possibility for even adaptive agents to get stuck in difficult policies.

The agent's actions optimized its own well-being and contributed to the resilience of the environment by preventing complete resource depletion. This reciprocal relationship between the agent and its environment is a fundamental aspect of sustainability, as the generative models of different layers in a hierarchical system are inherently linked through niche construction \cite{albarracin2024sustainability}.
However, the model's simplicity also reveals its limitations. The single-agent, single-resource setup does not capture the complex interdependencies and feedback loops present in real-world systems. Future research should explore multi-agent scenarios with competing interests and shared resources, as well as environments with multiple, interconnected resource types and more sophisticated replenishment dynamics.
Additionally, the model does not consider the possibility of permanent resource depletion, which would require conditioning the environment's survival on the maintenance of certain values. In the future, we need to incorporate this aspect to understand the long-term implications of resource management strategies.
To further advance the application of active inference in sustainable resource management, future work should focus on integrating network theory and dynamical systems theory to model and quantify the interdependencies between resources and their impact on overall system sustainability. Optimizing precision or learning rates could also help foster the elastic and plastic resilience necessary for long-term sustainability and abundance. We would need to explore this avenue further.
Our paper presents a proof-of-concept model demonstrating how active inference can inform sustainable resource management at the individual level. We consider the relationship between agent and environment to highlight the importance of resilience, adaptability, and long-term planning in achieving sustainable outcomes. While the model's simplicity limits its direct applicability to real-world systems, it provides a foundation for future research exploring the complex dynamics of sustainable resource management through the lens of active inference.

\bibliography{full_samplepaper}

\section{Appendix 1 - Figures}

\renewcommand{\figurename}{Fig.}

\renewcommand{\thefigure}{3}
\begin{figure}[htbp]
\centering
\includegraphics[width=\textwidth]{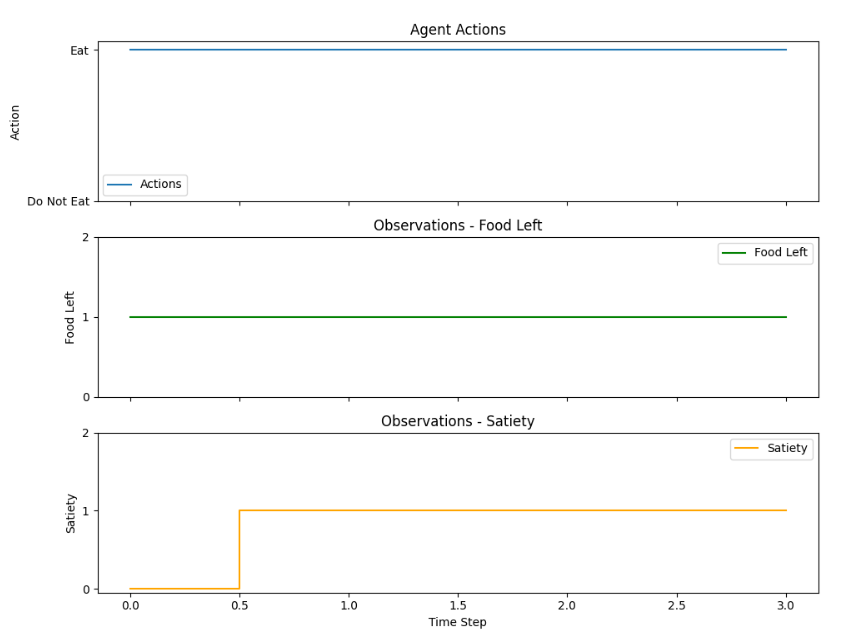}
\caption{Case 1. The three plots above show the expected behavior. At time step 0, food is available (food left = 0), and the satiety level is low (satiety = 0). Due to the agent's strong preference for high level of satiety, it keeps eating at subsequent time steps and the satiety increases. Since the environment is static, the food is always present.}
\label{fig:image14}
\end{figure}

\renewcommand{\thefigure}{4}

\begin{figure}[htbp]
\centering
\includegraphics[width=\textwidth]{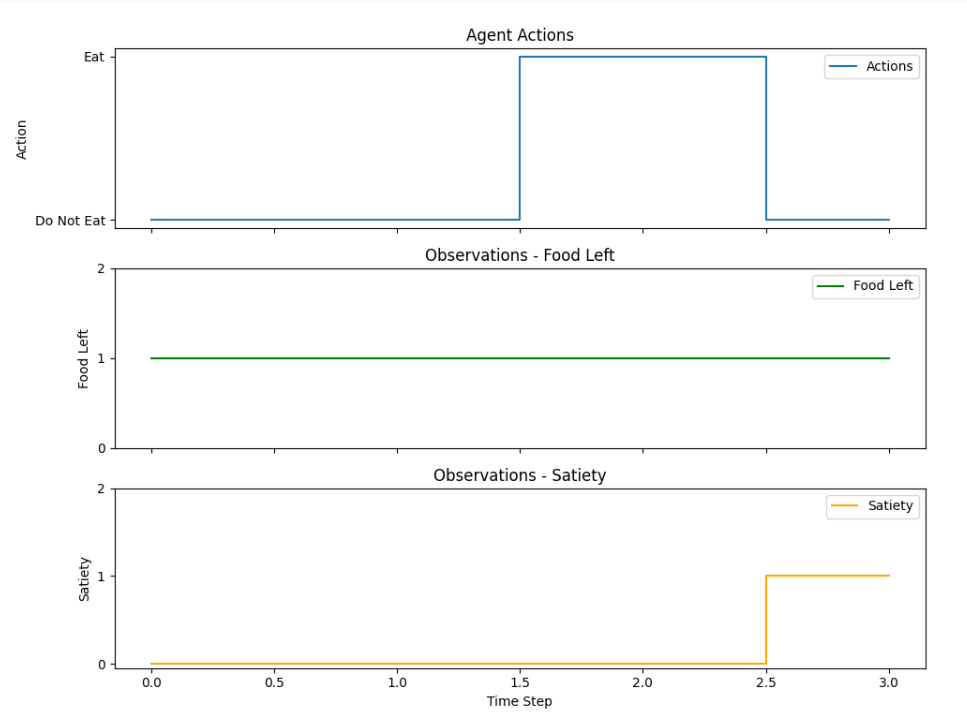}
\caption{Case 1.1 - where the agent is given incorrect A and B matrices, introducing errors in its perception and beliefs about state transitions.
The top plot shows the agent's actions over time. The pattern is more erratic compared to the standard Case 1, as the agent is confused.
The middle plot shows the food left observations. Food availability remains constant at 1 throughout the simulation since the environment is static.
The bottom plot shows the agent's satiety over time. Satiety level fluctuates more. This indicates that the agent's ability to maintain a stable, high satiety state is impaired by the incorrect perception and planning models.}

\label{fig:image4}
\end{figure}

\newpage

\renewcommand{\thefigure}{5}
\begin{figure}[htbp]
\centering
\begin{subfigure}{}
\includegraphics[width=.45\textwidth]{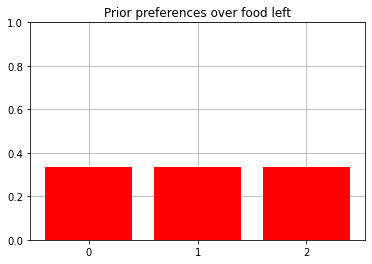}
\label{fig:image4a}
\end{subfigure}
\begin{subfigure}{}
\includegraphics[width=.45\textwidth]{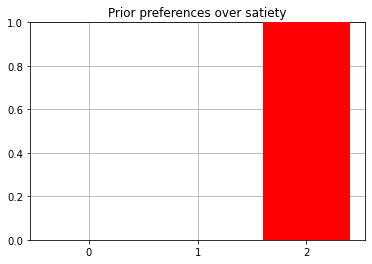}
\label{fig:image4b}
\end{subfigure}
\label{fig:image5}
\caption{Prior preference for Case 2. The  agent's preferences are changed so that, unlike case 1 and 1.1 it no longer has a preference over food left. Its only non-uniform preference is to have a preference over satiety.}
\end{figure}


%
\renewcommand{\thefigure}{6}
\begin{figure}[htbp]
  \centering
  \begin{subfigure}{}
  \includegraphics[width=.45\textwidth]{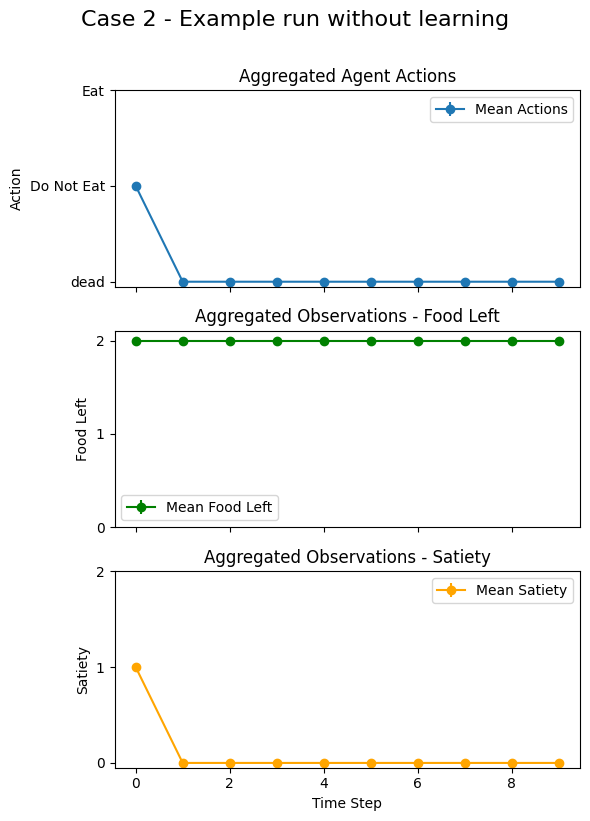}
  \end{subfigure}
  \begin{subfigure}{}
  \includegraphics[width=.45\textwidth]{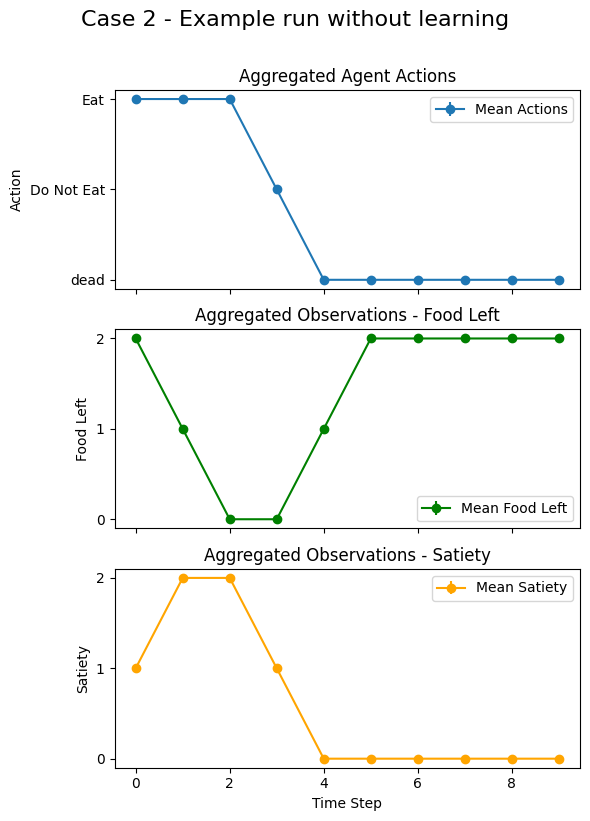}
  \end{subfigure}
  \caption{Case 2 - Dynamic Environment. Without learning, the agent either does not eat (as shown in the three plots on the left) or eats too much and therefore allows food in the environment to go to 0 (as shown in the three plots on the right). As a result, the agent dies.}
  \label{fig:image7}
\end{figure}

\renewcommand{\thefigure}{7}
\begin{figure}[htbp]
\centering
\includegraphics[width=\textwidth]{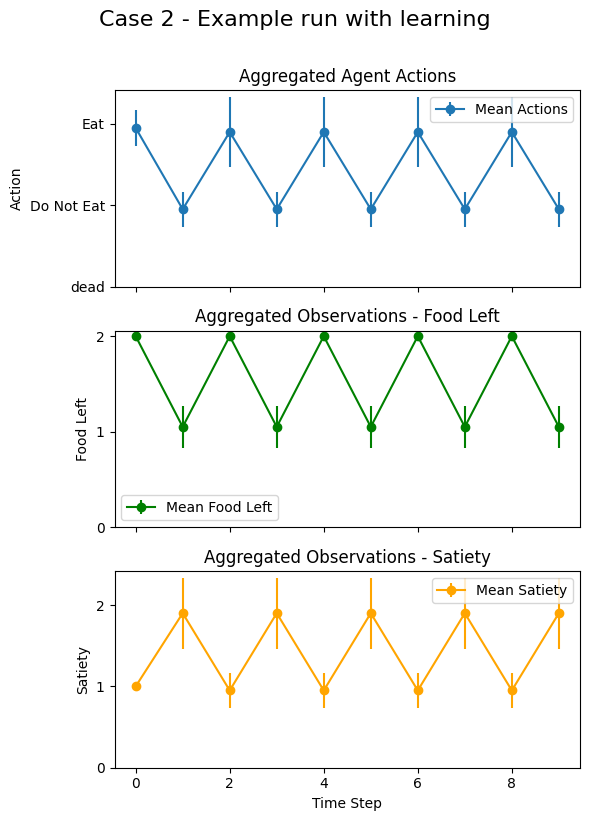}
\caption{Case 2 - Dynamic Environment. Example run with learning on and policy length = 3. With this depth of policy, the agent is able to plan further in time, and with learning it manages to survive for the whole length of the run. Over 10 time steps the agent is able to plan its behaviour so that it never reaches satiety = 0, and always has food left.}

\label{fig:image8}
\end{figure}



\renewcommand{\thefigure}{8}
\begin{figure}[htbp]
\centering
\hspace*{-0.25\textwidth}
\includegraphics[width=1.6\textwidth]{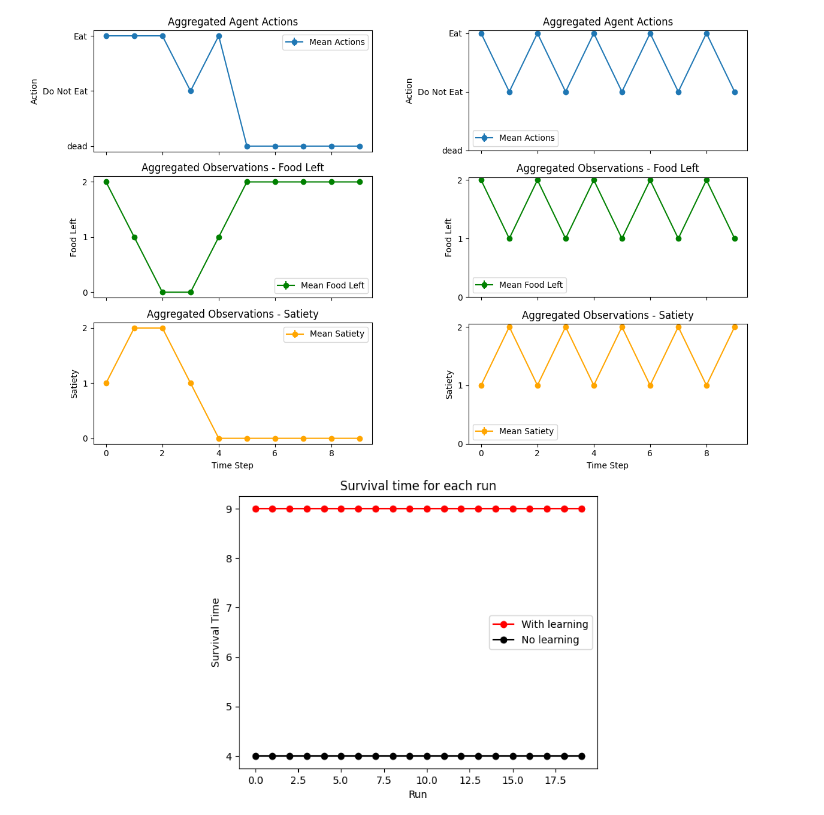}
\caption{Example run from Case 2 without learning enabled on the left and with learning enabled on the right, but starting with an extreme B matrix setting (probabilities set to 1 or 0, on the left three plots and middle three plots, and high but non-extreme values on the right). The agent dies quickly, just as the randomly set values of the B matrix in plot 6, and is able to learn on the right.}

\label{fig:image10}
\end{figure}

\renewcommand{\thefigure}{9}
\begin{figure}[htbp]
\centering
\hspace*{-0.25\textwidth}
\includegraphics[width=1.6\textwidth]{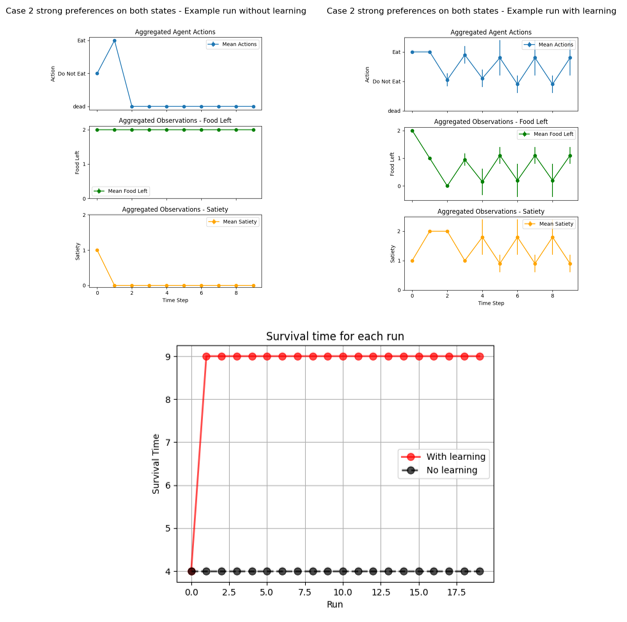}
\caption{Case 2 with strong prior preferences on both high satiety and high food left states. On the three top left plots, the agent has no learning, and on the top right, the agent has learning. On the bottom, we can see that survival time is vastly different with and without learning, as the preferences affect the behavior of the agent.}

\label{fig:image12}
\end{figure}

\renewcommand{\thefigure}{10}
\begin{figure}[htbp]
\centering
\hspace*{-0.3\textwidth}
\includegraphics[width=1.6\textwidth]{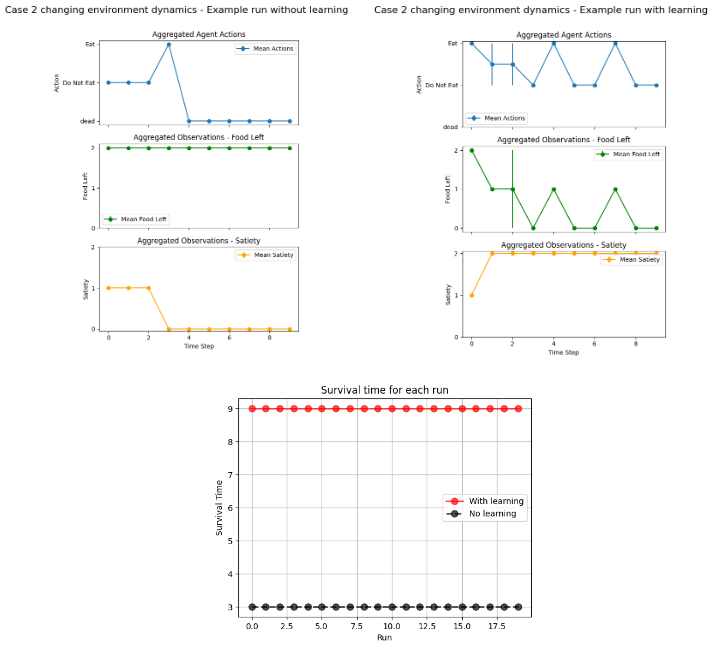}\caption{Case 2 in a changing environment where food and satiety change at different time rates. The three top left plots show the results without learning off, and the three top right plots show the results with learning on. The bottom plot represents the comparison between the survival time over 10 time steps.
Food increases at a slower rate (0.5 units per step) when not eating and decreases at a faster rate (1 unit per step) when eating. Satiety decreases faster when not eating (0.2 units per step) and increases at a different rate when eating (0.8 units per step). 
}
\label{fig:image13}
\end{figure}

\renewcommand{\thefigure}{11}
\begin{figure}[htbp]
\centering
\hspace*{-0.2\textwidth}
\includegraphics[width=1.4\textwidth]{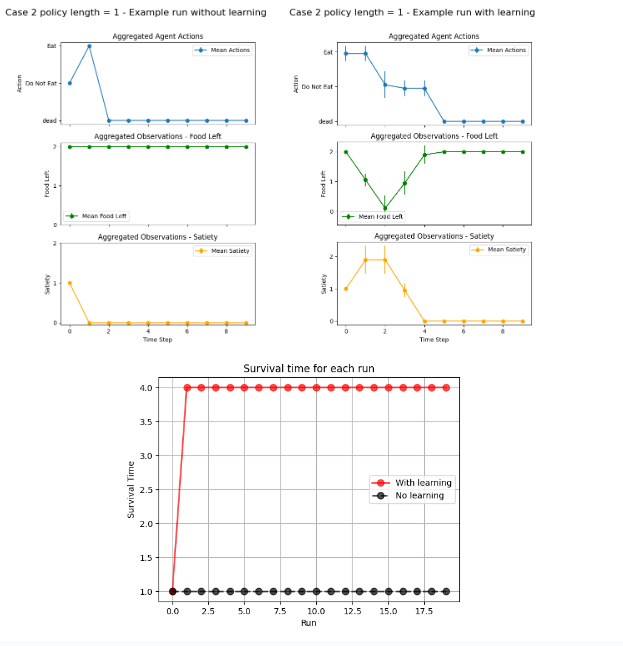}\caption{Case 2 example runs runs with policy length = 1, left plot without learning, right plot with learning,  and survival time on the bottom.}

\label{fig:image13}
\end{figure}

\end{document}